\documentclass[10pt]{article} 
\usepackage[preprint]{tmlr}

\usepackage{amsmath,amsfonts,bm}

\def\eqref#1{equation~\ref{#1}}

\def\1{\bm{1}}

\DeclareMathAlphabet{\mathsfit}{\encodingdefault}{\sfdefault}{m}{sl}
\SetMathAlphabet{\mathsfit}{bold}{\encodingdefault}{\sfdefault}{bx}{n}

\usepackage{hyperref}
\usepackage{url}
\usepackage{graphicx}
\usepackage{amsmath}
\usepackage{amsthm}
\usepackage{amssymb}
\usepackage{mathtools}
\usepackage{booktabs}
\usepackage{float}
\usepackage{placeins}

\theoremstyle{plain}

\title{Self-Distillation as a Performance Recovery Mechanism for LLMs: Counteracting Compression and Catastrophic Forgetting}

\author{Chi Liu \quad Xin Chen \quad Xu Zhou \quad Fangbo Tu \quad Srinivasan Manoharan \\
      {\normalfont\normalsize PayPal AI}}

\def\month{MM}  
\def\year{YYYY} 
\def\openreview{\url{https://openreview.net/forum?id=XXXX}} 

\begin{document}

\maketitle

\begin{abstract}
Large Language Models (LLMs) suffer from performance degradation across multiple scenarios: catastrophic forgetting during supervised fine-tuning, quantization, and pruning. Self-Distillation Fine-Tuning (SDFT) has been proposed as a method to restore degraded performance, but its effectiveness across different degradation types and its underlying mechanism remain to be systematically examined. In this work, we first establish that SDFT is a broadly effective and robust recovery method: across forgetting, quantization, and pruning, SDFT consistently recovers performance, supported by extensive baseline comparisons, multi-seed experiments, and long task sequences. We then investigate whether this recovery can be explained by a unified geometric mechanism --- specifically, the hypothesis that SDFT operates through ``manifold realignment'' (the student's hidden representations moving toward the teacher's), quantified by Centered Kernel Alignment (CKA). Our results reveal a differentiated picture: a consistent positive association between CKA increase and performance recovery is observed in both quantization and forgetting, supported by causal intervention experiments showing that suppressing CKA weakens recovery at the aggregate level. In pruning, however, we observe clear counterexamples --- SDFT recovers performance while CKA decreases or stagnates --- indicating that manifold alignment is not a universal requirement for recovery. These findings demonstrate that SDFT is an effective and robust recovery method across diverse degradation scenarios, while the manifold alignment hypothesis holds in quantization and forgetting but fails in pruning. Our work provides both a solid empirical foundation for SDFT and a boundary condition on the geometric interpretation of its recovery mechanism.
\end{abstract}

\section{Introduction}
\label{sec:introduction}

Large Language Models (LLMs) have revolutionized natural language understanding, reasoning, and generation. However, deploying generic base models into real-world applications necessitates further adaptation. To align with specific downstream tasks, models typically undergo Supervised Fine-Tuning (SFT); simultaneously, to meet resource constraints, techniques such as pruning and quantization become indispensable.

However, these operations often incur significant performance degradation. In continual learning, multi-round SFT frequently triggers Catastrophic Forgetting, where models lose original general knowledge and skills while acquiring new domain-specific knowledge and task capabilities. Similarly, aggressive compression disrupts internal parameter distributions, leading to declines in accuracy and logical consistency. This ``capability trade-off'' forces difficult choices between specialization and generalization. Once a model degrades, traditional repair methods are often computationally prohibitive, sometimes requiring retraining from scratch --- a highly inefficient solution in the context of scarce computational resources.

Recent work has explored Self-Distillation Fine-Tuning (SDFT) as a method for mitigating forgetting in continual learning \citep{shenfeld2026sdft} and for recovering quality in pruned models \citep{thangarasa2025self}. These studies have demonstrated the empirical effectiveness of SDFT in isolated settings. However, three key questions remain unanswered: (1) Does SDFT generalize as an effective recovery method across a broader range of degradation scenarios? (2) Is SDFT's recovery effect robust under different initializations and more complex task sequences? (3) What is the underlying mechanism --- is there a unified explanation for why SDFT works, or do different scenarios operate through different mechanisms?

In this work, we address these questions through a systematic empirical investigation. We first establish that SDFT is a broadly effective and robust recovery method. Across three degradation scenarios (catastrophic forgetting, quantization, and pruning) SDFT consistently recovers model performance. We validate this effectiveness through extensive baseline comparisons against standard SFT, knowledge distillation, and replay-based methods, across multiple random seeds, and on longer, more realistic task sequences.

We then investigate the mechanistic question. A prominent hypothesis in the literature is that SDFT operates through ``manifold realignment'' --- the idea that the student model's hidden representations move toward the teacher's, restoring the geometric structure of the representation space. To test this hypothesis, we employ Centered Kernel Alignment (CKA) \citep{kornblith2019similarity} as a quantitative measure of representation similarity between the student and teacher models.

Our results reveal a differentiated picture. We find a consistent positive association between CKA increase and performance recovery in the quantization and catastrophic forgetting scenarios. In pruning, we observe no such consistent relationship: recovery can occur without CKA increase, and CKA can decrease while performance improves. This finding does not undermine the empirical value of SDFT --- it remains an effective recovery method in all three scenarios --- but it does provide a boundary condition on the manifold alignment hypothesis, suggesting that SDFT's recovery mechanism is more diverse than a single geometric principle.

The remainder of this paper is organized as follows. Section~\ref{sec:related_work} discusses related work. Section~\ref{sec:recovery_framework} describes the unified SDFT recovery framework for different degradation scenarios. Section~\ref{sec:manifold_hypothesis} formalizes the manifold alignment hypothesis and introduces CKA as a quantitative measure. Section~\ref{sec:experiments} presents experimental results, including baseline comparisons, multi-seed validation, long task sequences, and the CKA analysis across all three scenarios. Section~\ref{sec:discussion} discusses implications and limitations. Section~\ref{sec:conclusion} concludes.

\section{Related Work}
\label{sec:related_work}

\paragraph{Catastrophic Forgetting in LLMs.}
Catastrophic forgetting (CF) refers to the phenomenon wherein neural networks suddenly and significantly lose previously learned knowledge when trained on new data \citep{de2021continual}. In the context of LLMs, this manifests when multi-round SFT overwrites knowledge acquired in prior trainings \citep{li2017learning}. Existing mitigation strategies fall into several categories: (1) replay-based methods, which store a subset of old data to interleave with new training \citep{de2021continual}; (2) regularization-based methods, such as Elastic Weight Consolidation (EWC), which penalize changes to important parameters \citep{kirkpatrick2017overcoming}; (3) parameter-isolation methods, which allocate separate parameters for different tasks \citep{rusu2016progressive}; and (4) distillation-based methods, such as Learning without Forgetting (LwF) \citep{li2017learning}, which use the pre-forgetting model to regularize the output distribution during new task training. While effective, these approaches primarily focus on preventing forgetting during the training phase itself. They often incur high computational costs, require active access to historical data or models throughout the SFT phase, or complicate model architecture. Crucially, our framework addresses a different phase of the problem: we focus on post-hoc recovery, restoring degraded capabilities after the initial SFT or compression has already occurred.

\paragraph{Model Compression and Performance Degradation.}
To deploy LLMs efficiently, techniques such as pruning \citep{ma2024llmpruner} and quantization \citep{dettmers2023qlora} are widely adopted. However, these operations inevitably introduce performance degradation \citep{frantar2022gptq}. Traditional remedies often rely on Knowledge Distillation (KD), where a compressed student model is trained to mimic a larger, uncompressed teacher \citep{hinton2015distilling}. While external strong teachers are theoretically applicable, they often introduce distribution shifts, high computational overhead, or privacy constraints. Prior work \citep{dettmers2023qlora, ma2024llmpruner} has demonstrated that post-compression fine-tuning can recover strong performance. However, these studies treated recovery as an empirical observation without providing a mechanistic interpretation of why additional fine-tuning restores performance at the representation level --- a gap our work aims to address.

\paragraph{Self-Distillation Fine-Tuning.}
Self-distillation (SD) has emerged as a technique for enhancing model generalization without relying on an external teacher. Early works demonstrated that training a model to mimic its own deeper layers or earlier checkpoints acts as an effective regularizer \citep{furlanello2018born}. More recently, \citet{shenfeld2026sdft} introduced Self-Distillation Fine-Tuning (SDFT) for mitigating catastrophic forgetting in continual learning, where the model uses its current state as a teacher to generate on-policy training signals. Separately, \citet{thangarasa2025self} explored self-data distillation for recovering quality in pruned LLMs, using the original unpruned model as a teacher to generate distillation data for fine-tuning the pruned model.

\paragraph{Our work in context.}
We acknowledge the methodological similarity to \citet{shenfeld2026sdft} and \citet{thangarasa2025self} --- we also employ SDFT as a recovery mechanism. Our contribution differs in two respects. First, we systematically evaluate SDFT across three degradation scenarios (forgetting, quantization, pruning) with extensive baseline comparisons, multi-seed validation, and long task sequences, establishing its effectiveness and robustness. Second, we test the manifold alignment hypothesis as a potential explanation for SDFT's effectiveness, finding that it holds in the quantization and forgetting scenarios, thereby providing a boundary condition on the geometric interpretation of SDFT's recovery mechanism.

\section{Recovery via Self-Distillation Fine-Tuning}
\label{sec:recovery_framework}

\subsection{Formalizing the Degradation and Recovery Process}

Let $M_\theta$ denote a Large Language Model parameterized by $\theta$. In practical deployment, $M_\theta$ often undergoes a degradation operator $D$, resulting in a degraded model $M_{\theta'} = D(M_\theta)$ that exhibits significant performance drops in previously acquired capabilities.

Our objective is to investigate the post-hoc recovery of $M_{\theta'}$. We employ Self-Distillation Fine-Tuning (SDFT) \citep{shenfeld2026sdft} as the recovery mechanism across all scenarios. Following the SDFT formulation, we define a reference model $M_{\mathrm{ref}}$ (which retains the target capabilities) and optimize the degraded model (acting as the student $M_{\mathrm{stu}}$, initialized from $M_{\theta'}$) to mimic the output distribution of $M_{\mathrm{ref}}$:
\begin{equation}
\mathcal{L}_{\mathrm{SDFT}} = \mathbb{E}_{x \sim X}\left[ \mathrm{KL}\!\left( P(y \mid x; M_{\mathrm{ref}}) \,\|\, P(y \mid x; M_{\mathrm{stu}}) \right) \right]
\end{equation}

By maintaining this identical optimization objective across diverse degradation operators $D$, we establish a controlled setting to examine whether any consistent geometric pattern quantified by CKA accompanies recovery, and to assess its generality across different degradation types. We now describe how this SDFT setup is instantiated across the three scenarios.

\subsection{Instantiation Across Degradation Scenarios}

\begin{figure}[htbp]
\centering
\includegraphics[width=\textwidth]{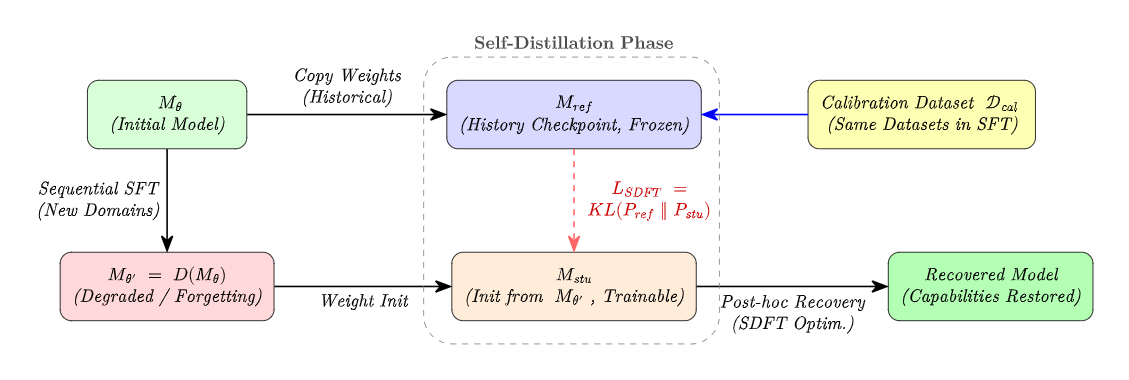}
\caption{The SDFT recovery framework for catastrophic forgetting.}
\label{fig:figure1}
\vspace{-1ex}
\end{figure}

\paragraph{Catastrophic Forgetting (Figure~\ref{fig:figure1}).}
In the continual learning scenario, the degradation operator $D$ represents sequential SFT on new domains, causing $M_{\theta'}$ to forget prior knowledge. The reference model $M_{\mathrm{ref}}$ is instantiated as the historical checkpoint $M_\theta$ prior to the new task adaptation. The SDFT process enables the student to recover forgotten capabilities by aligning with its own historical state, without requiring external high-performance teachers or proprietary datasets.

\begin{figure}[htbp]
\centering
\includegraphics[width=\textwidth]{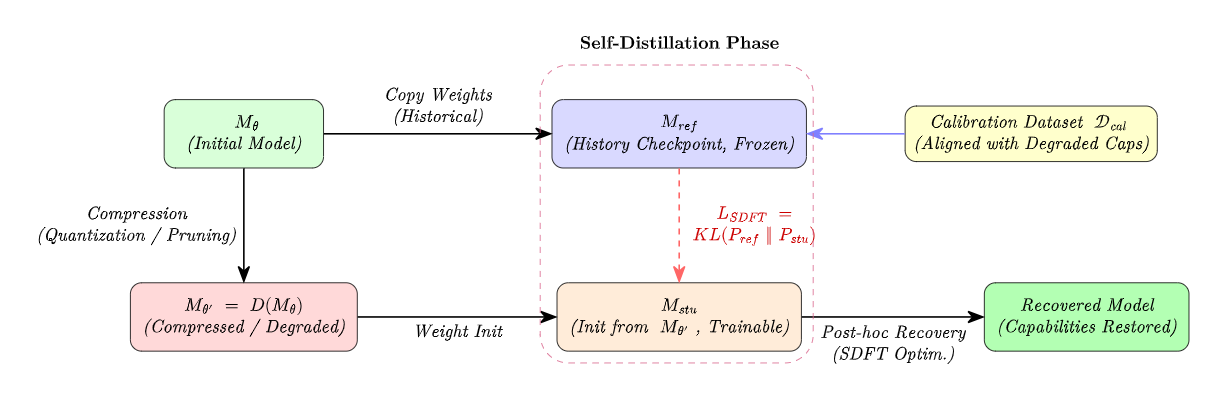}
\caption{The SDFT recovery framework for model compression (quantization and pruning).}
\label{fig:figure2}
\vspace{-1ex}
\end{figure}

\paragraph{Model Compression (Figure~\ref{fig:figure2}).}
When $D$ represents compression (quantization and pruning), $M_{\theta'}$ suffers from performance degradation. Here, $M_{\mathrm{ref}}$ is the original model $M_\theta$. To facilitate the distillation process, the calibration datasets aligned with the degraded capabilities are utilized. Despite the differing physical nature of the degradation, the underlying SDFT recovery objective remains identical.

\subsection{Extension to Small-Scale Models}

While the SDFT formulation described above is effective for medium-to-large scale models, its efficacy diminishes at smaller scales (e.g., $\le 3$B parameters). This limitation stems from the fact that SDFT relies heavily on the model's intrinsic In-Context Learning (ICL) capabilities to provide meaningful teacher guidance, which are typically underdeveloped in smaller models.

\begin{figure}[htbp]
\centering
\includegraphics[width=\textwidth]{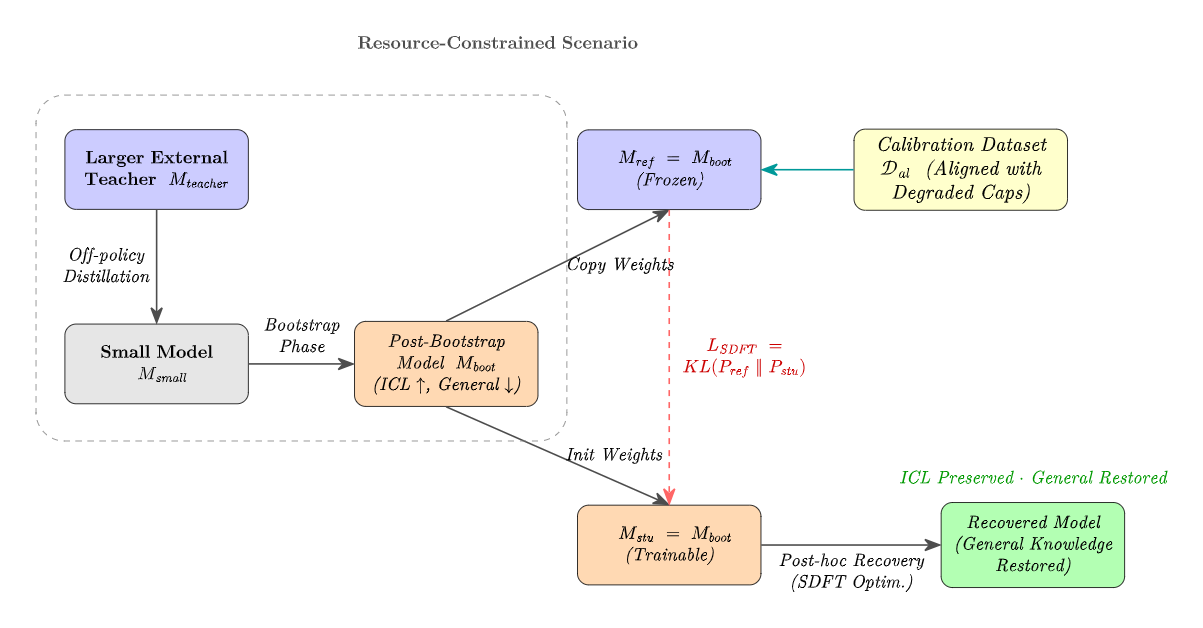}
\caption{Extension to small-scale models: a bootstrap phase (off-policy distillation from a larger teacher) followed by SDFT recovery.}
\label{fig:figure3}
\vspace{-1ex}
\end{figure}

To extend SDFT recovery to resource-constrained scenarios, we introduce a preliminary bootstrap phase (Figure~\ref{fig:figure3}). We first employ off-policy distillation using a larger external teacher to enhance the small model's ICL capabilities. While this step improves ICL, it can induce degradation in general domain knowledge as a side effect. Subsequently, we apply the same SDFT recovery mechanism (using the post-bootstrap model as $M_{\mathrm{ref}}$) to restore these degraded capabilities.

Crucially, for the purpose of examining the manifold alignment hypothesis, we focus on the realignment that occurs during the SDFT recovery phase, not on the bootstrap step. The external teacher is utilized strictly as a one-time step to satisfy the ICL prerequisite, preserving the self-contained nature of the core recovery mechanism we analyze.

\section{The Manifold Alignment Hypothesis}
\label{sec:manifold_hypothesis}

\subsection{Introducing the Hypothesis}

A natural hypothesis for why SDFT recovers performance is that it operates through manifold realignment --- the idea that the student model's hidden representations move toward the teacher's, thereby restoring the geometric structure of the representation space. Under this hypothesis, the effectiveness of SDFT across different degradation scenarios (forgetting, quantization, and pruning) would be explained by a shared geometric mechanism: in all cases, recovery is accompanied by the student's manifold moving closer to the teacher's.

To test this hypothesis, we need a quantitative measure of manifold alignment. We employ Centered Kernel Alignment (CKA) \citep{kornblith2019similarity}, which measures the similarity between two representations in a way that is invariant to orthogonal transformations and scaling. We compute CKA between the student model and the teacher reference at the last hidden layer, using a fixed evaluation set.

If the manifold alignment hypothesis holds, we would expect to observe a consistent positive association between the change in CKA ($\Delta$CKA) and performance recovery ($\Delta$Acc) across all degradation scenarios. We test this prediction in Section~\ref{sec:experiments}. As we will show, the results reveal a differentiated picture: the association holds in quantization, but not in forgetting or pruning --- providing a boundary condition on the hypothesis.

\subsection{Formalization: Activation Trajectories and Manifold Geometry}

To test the manifold alignment hypothesis, we need a formal way to represent and compare the geometric structure of hidden representations. This section introduces the necessary notation and definitions.

Given an input sequence $X = (x_1, x_2, \ldots, x_L)$, where $L$ denotes the sequence length. For a certain hidden layer of an LLM (e.g., the last hidden layer), each token $x_t$ corresponds to a $d$-dimensional activation vector $h_t \in \mathbb{R}^d$. We stack the activation vectors of all tokens from a complete forward pass to form the Activation Matrix $H \in \mathbb{R}^{L \times d}$:
\begin{equation}
H = [\,h_1, h_2, \ldots, h_L\,]^\top
\end{equation}

From the perspective of manifold learning, each row in $H$ represents a sample point on the high-dimensional semantic manifold $\mathcal{M}$, and the entire matrix $H$ constitutes a discrete trajectory of the sequence on this manifold. The student model $S$ and the teacher model $T$ generate activation matrices $H_S$ and $H_T$, respectively. Our objective is to measure the geometric alignment between these two trajectories. This measurement allows us to quantify whether the student's representation structure is moving toward the teacher's during recovery, and to what extent.

We do not claim to characterize the complete underlying manifolds. Instead, we utilize activation trajectories as discrete samples, which provide a computable and tractable basis for comparing the geometric structure of different model states.

Directly comparing the element values of activation matrices $H_S$ and $H_T$ (e.g., using MSE) is inappropriate because neural network representations possess rotation invariance. Semantically identical features may exist along different coordinate axes in the hidden space. To capture the geometric structure of the manifold, we need to measure the relative relationships between tokens rather than their absolute coordinates.

We compute the Linear Kernel Matrix $K \in \mathbb{R}^{L \times L}$:
\begin{equation}
K = H H^\top
\end{equation}

Here, the element $K_{ij} = h_i \cdot h_j$ represents the similarity between the $i$-th and $j$-th tokens in the hidden space. The matrix $K$ thus reflects the pairwise relational structure within the sequence and serves as a tractable proxy for the geometric properties of the manifold.

\subsection{Quantifying Manifold Alignment via CKA}

To quantify the degree of manifold alignment between the student and teacher models, we adopt Centered Kernel Alignment (CKA) \citep{kornblith2019similarity}. The following six steps describe the calculation procedure.
\begin{enumerate}
\item \textbf{Input Consistency.} Pass the identical input sequence (Prompt + Ground Truth) into both the student and teacher models to ensure token-level correspondence.
\item \textbf{Activation Extraction.} Extract the same-layer activation matrices $H_S, H_T \in \mathbb{R}^{L \times d}$.
\item \textbf{Kernel Matrix Computation.} Compute the linear kernel matrices $K_S = H_S H_S^\top$ and $K_T = H_T H_T^\top$.
\item \textbf{Centering Operation.} Construct the centering matrix $C = I_L - \frac{1}{L}\mathbf{1}\mathbf{1}^\top$. Compute the centered kernel matrices $\tilde{K}_S = C K_S C$ and $\tilde{K}_T = C K_T C$. This step removes global biases in activation values, ensuring the metric reflects relative structure rather than absolute coordinates.
\item \textbf{Hilbert-Schmidt Independence Criterion (HSIC) \citep{gretton2005measuring}.} Compute the Frobenius inner product of the two centered kernel matrices:
\begin{equation}
\mathrm{HSIC}(H_S, H_T) = \mathrm{tr}\!\left( \tilde{K}_S \tilde{K}_T \right)
\end{equation}
The original HSIC definition includes a scaling factor $\frac{1}{(L-1)^2}$ for unbiased estimation; this factor cancels out in the normalized CKA ratio and is thus omitted for brevity.
\item \textbf{CKA Normalization.} The final alignment score is calculated as:
\begin{equation}
\mathrm{CKA}(H_S, H_T) = \frac{\mathrm{HSIC}(H_S, H_T)}{\sqrt{\mathrm{HSIC}(H_S, H_S) \cdot \mathrm{HSIC}(H_T, H_T)}}
\end{equation}
\end{enumerate}

The value of CKA ranges from 0 to 1. A score closer to 1 indicates higher geometric similarity between the student's and teacher's activation trajectories. We use CKA to track this similarity over the course of recovery --- specifically, to observe whether the student's manifold moves toward the teacher's manifold.

\subsection{Framework and Experimental Design}

\paragraph{Computation and interpretation.}
We compute CKA between the student and the teacher on a fixed evaluation set during recovery. An increase in CKA ($\Delta$CKA $> 0$) indicates that the student's manifold is moving toward the teacher's manifold; a decrease suggests further drift.

\paragraph{Limitations.}
CKA values are computed from activations produced by a specific input dataset. Consequently, absolute CKA values are not comparable across different datasets. Our analysis relies on relative changes ($\Delta$CKA) within a fixed dataset throughout the recovery process, i.e., comparing the CKA values computed between the student and the teacher at different stages of recovery for the same task.

\paragraph{Experimental design.}
To assess the manifold alignment hypothesis, we conduct two sets of experiments. First, we test whether a positive association between $\Delta$CKA and performance recovery exists across the three degradation scenarios, i.e., does recovery consistently accompany manifold realignment? Second, we conduct a causal intervention experiment that penalizes CKA increase during SDFT recovery (i.e., we intentionally discourage the student's manifold from moving toward the teacher's) to probe whether manifold realignment plays a functional role. As we will show in Section~\ref{sec:experiments}, both tests yield consistent results: the $\Delta$CKA--$\Delta$Acc association and the causal intervention effect are only observed in the quantization scenario, but not in forgetting or pruning. These findings provide a boundary condition on the manifold alignment hypothesis.

\section{Experiments}
\label{sec:experiments}

\subsection{Experimental Setup [TODO: Need confirmation]}
\label{sec:exp_setup}

\paragraph{Models.} Qwen2.5-3B-Instruct \citep{hui2024qwen2} or Qwen3.5-9B \citep{qwen2026qwen35} or Gemma-4-12B \citep{gemmateam2026gemma4} used in different experiments.

\paragraph{Tasks.} Science, Tooluse, GSM8K (math), ARC-Challenge (reasoning), HellaSwag (commonsense), TruthfulQA (factuality). General capability: MMLU and Winogrande.

\paragraph{Degradation.} Three scenarios: (1) catastrophic forgetting via SFT (LoRA or full); (2) NF4 quantization (with 8-bit GPTQ variant); (3) Wanda structured pruning at 40\% and 50\%.

\paragraph{Baselines.} Standard SFT, forward-KL KD \citep{hinton2015distilling}, EWC \citep{kirkpatrick2017overcoming}, LwF \citep{li2017learning}, and Replay (joint training).

\paragraph{Implementation (for all if not specified).} LoRA: r=16, alpha=32, dropout=0.05, targets q/k/v/o/gate/up/down. LR=5e-5, cosine with 10\% warmup, effective batch 32, 2 epochs. SDFT uses on-policy reverse KL (alpha=1); teacher = bf16 (compression) or pre-degradation checkpoint (forgetting). For compression recovery, each method is trained independently on Tooluse and Science (separate runs).

\paragraph{CKA.} Computed at the last hidden layer on a fixed eval set if not specified, using linear kernel and global centering.

\subsection{SDFT is Effective and Robust}
\label{sec:sdft_effective}

We first establish whether SDFT provides meaningful recovery compared to alternative methods across different degradation scenarios. We evaluate SDFT on three settings: NF4 quantization, Wanda 50\% pruning, and catastrophic forgetting.

\paragraph{Quantization Recovery (NF4 4-bit).}
Using Gemma-4-12B under NF4 quantization, we compare SDFT against SFT (LoRA) and forward-KL KD. For each method, we run independent recoveries on Science and Tooluse separately. We evaluate task-specific recovery, and general capability across six benchmarks (MMLU, Winogrande, ARC-C, HellaSwag, GSM8K, TruthfulQA).

\begin{table}[htbp]
\centering
\small
\resizebox{\textwidth}{!}{%
\begin{tabular}{@{}lllcccccc@{}}
\toprule
Method & Task & Task Acc\% ($\Delta$ vs.\ 4bit) & MMLU & Wino & ARC-C & HellaS & GSM8K & TruQA \\
\midrule
bf16 (reference) & Tooluse \& Science & 51.55 (Tool) / 52.66 (Sci) & 77.6 & 70.6 & 49.5 & 58.7 & 86.7 & 62.5 \\
NF4 4bit & Tooluse \& Science & 40.21 (Tool) / 51.08 (Sci) & 74.9 & 68.8 & 47.9 & 60.5 & 84.1 & 60.7 \\
SDFT & Tooluse & 52.58 (+12.37) & 75.7 & 67.9 & 48.6 & 60.9 & 85.5 & 59.3 \\
SDFT & Science & 59.17 (+8.09) & 75.8 & 69.0 & 47.4 & 61.2 & 84.7 & 59.6 \\
SFT (LoRA) & Tooluse & 67.01 (+26.80) & 74.0 & 71.3 & 53.5 & 66.4 & 85.3 & 55.6 \\
SFT (LoRA) & Science & 62.33 (+11.25) & 74.8 & 72.6 & 52.6 & 66.1 & 80.9 & 61.6 \\
Fwd-KL KD & Tooluse & 52.58 (+12.37) & 75.5 & 71.0 & 50.1 & 62.7 & 85.8 & 58.8 \\
Fwd-KL KD & Science & 53.85 (+2.77) & 76.2 & 70.0 & 49.5 & 62.7 & 84.8 & 59.9 \\
\bottomrule
\end{tabular}%
}
\caption{Baseline comparison on compression recovery (NF4 quantization) using Gemma-4-12B.}
\label{tab:quant_baseline}
\end{table}

\paragraph{Pruning Recovery (Wanda 50\% Structured Pruning).}
We further evaluate SDFT on a more challenging compression setting: Wanda 50\% structured pruning on Gemma-4-12B. In addition to SFT (LoRA), we include EWC and LWF as additional baselines. General capability is measured via MMLU, additional benchmarks are left for future work.

\begin{table}[htbp]
\centering
\small
\begin{tabular}{@{}llcc@{}}
\toprule
Method & Task & Task Acc\% ($\Delta$ vs.\ Pruned Base) & MMLU \\
\midrule
bf16 (reference) & Tooluse \& Science & 51.55 (Tool) / 52.66 (Sci) & 77.6 \\
Pruned base & Tooluse \& Science & 51.55 (Tool) / 40.24 (Sci) & 55.21 \\
SDFT & Tooluse & 60.82 (+9.27) & 62.73 \\
SDFT & Science & 46.55 (+6.31) & 60.13 \\
SFT (LoRA) & Tooluse & 65.98 (+14.43) & 53.99 \\
SFT (LoRA) & Science & 53.65 (+13.41) & 60.35 \\
EWC (anchor=Pruned base, $\lambda$=1e4) & Tooluse & 68.04 (+16.49) & 57.26 \\
EWC (anchor=Pruned base, $\lambda$=1e4) & Science & 44.77 (+4.53) & 56.69 \\
LWF (anchor=Pruned base, $\lambda$=1.0) & Tooluse & 62.89 (+11.34) & 55.97 \\
LWF (anchor=Pruned base, $\lambda$=1.0) & Science & 45.56 (+5.32) & 57.44 \\
\bottomrule
\end{tabular}
\caption{Baseline comparison on compression recovery (Wanda 50\% Pruning) using Gemma-4-12B.}
\label{tab:prune_baseline}
\end{table}

\paragraph{Catastrophic Forgetting Recovery.}
Finally, we evaluate SDFT in the catastrophic forgetting scenario. Using Qwen2.5-3B-Instruct as the initial model, we sequentially fine-tune it on Science and Tooluse tasks, and subsequently apply SDFT to recover Science knowledge using different teacher models.

\begin{table}[htbp]
\centering
\small
\begin{tabular}{@{}lcc@{}}
\toprule
Method & Tooluse Acc\% ($\Delta$ vs.\ Sci Expert) & Science Acc\% ($\Delta$ vs.\ Sci Expert) \\
\midrule
Science expert & 19.59 & 59.96 \\
Forgotten & 64.95 (+45.36) & 37.87 ($-$22.09) \\
SDFT (teacher=base) & 65.98 (+46.39) & 61.54 (+1.58) \\
SDFT (teacher=expert) & 57.73 (+38.14) & 65.48 (+5.52) \\
\bottomrule
\end{tabular}
\caption{SDFT recovery on catastrophic forgetting using Qwen2.5-3B-Instruct.}
\label{tab:forget_recovery}
\end{table}

\paragraph{Results.}
In the quantization recovery scenario (Table~\ref{tab:quant_baseline}), SDFT achieves substantial recovery on both Tooluse (+12.37) and Science (+8.09) from the NF4 baseline. Compared to SFT (LoRA) and fwd-KL KD, SDFT performs competitively while maintaining general capabilities across six benchmarks.

In the pruning recovery scenario (Table~\ref{tab:prune_baseline}), pruning leaves Tooluse unaffected (51.55 $\to$ 51.55) while substantially degrading Science (52.66 $\to$ 40.24, $-$12.42). All recovery methods substantially improve Tooluse beyond the original bf16 level, indicating that these gains are due to task-specific training rather than recovery of lost capability. SDFT recovers Science by +6.31 and increases Tooluse by +9.27 from the pruned baseline. SFT achieves higher task-specific recovery on Science (+13.41) but yields lowest MMLU (53.99) among all methods in this scenario, EWC and LWF show competitive Tooluse increase but substantially weaker Science recovery (44.77 and 45.56, respectively) compared to SDFT (46.55).

In the forgetting recovery scenario (Table~\ref{tab:forget_recovery}), SDFT effectively mitigates catastrophic forgetting. After sequential training on Tooluse, Science accuracy drops substantially from 59.96 to 37.87. Applying SDFT with the base model as teacher restores Science accuracy to 61.54 while maintaining Tooluse performance at 65.98. Using the Science expert as teacher achieves superior Science recovery (65.48) but incurs a notable degradation in Tooluse accuracy (57.73), revealing a trade-off between knowledge restoration and new capability retention.

\paragraph{Robustness Across Random Seeds.}
Having established SDFT recovery effectiveness in multiple scenarios, we next verify that this recovery pattern is not an artifact of a particular random seed. We repeat the forgetting-recovery pipeline on Qwen3.5-9B, using the same pipeline (Science $\to$ Tooluse $\to$ Recovery) as the catastrophic forgetting scenario, to assess SDFT's statistical robustness across three random seeds (42, 43, 44). We compare SDFT against SFT (LoRA), fwd-KL KD, and Replay, evaluating Science recovery, Tooluse retention, and general capability across six benchmarks (MMLU, Winogrande, ARC-C, HellaSwag, GSM8K, TruthfulQA).

\begin{table}[htbp]
\centering
\small
\resizebox{\textwidth}{!}{%
\begin{tabular}{@{}lccl@{}}
\toprule
Method & Science Acc\% (mean $\pm$ std) & $\Delta$ vs.\ Forgotten & Per-seed \\
\midrule
Science expert & 60.75 $\pm$ 0.32 & --- & s42=60.75 s43=61.14 s44=60.36 \\
Forgotten & 56.02 $\pm$ 2.95 & --- & s42=59.96 s43=55.23 s44=52.86 \\
SDFT & 62.52 $\pm$ 0.64 & +6.50 & s42=63.31 s43=62.52 s44=61.74 \\
SFT (LoRA) & 62.85 $\pm$ 1.13 & +6.83 & s42=61.54 s43=64.30 s44=62.72 \\
Fwd-KL KD & 60.62 $\pm$ 1.07 & +4.60 & s42=59.96 s43=59.76 s44=62.13 \\
Replay (needs old data) & 66.21 $\pm$ 1.03 & +10.19 & s42=65.68 s43=67.65 s44=65.29 \\
\bottomrule
\end{tabular}%
}
\caption{Science recovery with random seeds using Qwen3.5-9B.}
\label{tab:seed_science}
\end{table}

\begin{table}[htbp]
\centering
\small
\resizebox{\textwidth}{!}{%
\begin{tabular}{@{}lccl@{}}
\toprule
Method & Tooluse Acc\% (mean $\pm$ std) & $\Delta$ vs.\ Forgotten & Per-seed \\
\midrule
Science expert & 53.26 $\pm$ 0.97 & --- & s42=54.64 s43=52.58 s44=52.58 \\
Forgotten & 67.70 $\pm$ 1.28 & --- & s42=65.98 s43=68.04 s44=69.07 \\
SDFT & 68.73 $\pm$ 0.49 & +1.03 & s42=69.07 s43=68.04 s44=69.07 \\
SFT (LoRA) & 67.70 $\pm$ 1.28 & +0.00 & s42=65.98 s43=68.04 s44=69.07 \\
Fwd-KL KD & 68.38 $\pm$ 0.49 & +0.68 & s42=68.04 s43=68.04 s44=69.07 \\
Replay (needs old data) & 66.32 $\pm$ 0.49 & $-$1.38 & s42=65.98 s43=65.98 s44=67.01 \\
\bottomrule
\end{tabular}%
}
\caption{Tooluse retention after science recovery using Qwen3.5-9B.}
\label{tab:seed_tooluse}
\end{table}

\begin{table}[htbp]
\centering
\small
\begin{tabular}{@{}lcccccc@{}}
\toprule
Method & MMLU & Wino & ARC-C & HellaS & GSM8K & TruQA \\
\midrule
Science expert & 79.5 & 75.5 & 62.3 & 70.5 & 88.6 & 62.0 \\
Forgotten & 79.0 & 75.2 & 61.3 & 73.3 & 84.2 & 56.3 \\
SDFT & 79.2 & 75.3 & 62.2 & 73.1 & 87.0 & 58.5 \\
SFT (LoRA) & 79.5 & 75.3 & 63.1 & 73.2 & 87.6 & 59.0 \\
Fwd-KL KD & 79.2 & 76.3 & 62.8 & 73.4 & 87.0 & 58.7 \\
Replay (needs old data) & 79.4 & 75.4 & 62.5 & 73.4 & 86.7 & 58.0 \\
\bottomrule
\end{tabular}
\caption{General capability evaluation, four recovery methods have the same evaluation after science recovery, using Qwen3.5-9B.}
\label{tab:seed_general}
\end{table}

\paragraph{Results.}
SDFT exhibits the smallest standard deviation in Science recovery (0.64) among all recovery methods, compared to 1.13 for SFT (LoRA), 1.07 for Fwd-KL KD, and 1.03 for Replay. This confirms that SDFT's recovery effectiveness is robust across different initializations.

SDFT recovers Science to 62.52 $\pm$ 0.64 (+6.50 from forgotten) while perfectly preserving Tooluse at 68.73 $\pm$ 0.49 (+1.03). SFT (LoRA), while achieving slightly higher Science recovery (62.85 $\pm$ 1.13), shows larger variance and lower Tooluse retention (67.70 $\pm$ 1.28). Fwd-KL KD falls between the two, recovering less Science (60.62) with moderate Tooluse retention (68.38). Replay achieves the highest Science recovery (66.21) but requires old data and exhibits the worst Tooluse retention (66.32), highlighting the trade-off SDFT avoids.

SDFT maintains general capabilities across all six benchmarks, with values comparable to other methods. Notably, Science training and subsequent forgetting had heterogeneous effects on different capabilities: GSM8K (math) dropped from 88.6 to 84.2 ($-$4.4) and TruthfulQA (factuality) from 62.0 to 56.3 ($-$5.7), while HellaSwag (commonsense) actually increased from 70.5 to 73.3 (+2.8). SDFT partially recovers these degraded capabilities (GSM8K: 87.0, TruthfulQA: 58.5) without sacrificing others, confirming that the recovery does not come at the cost of broader capability degradation.

\paragraph{Long Task Sequence.}
Having established that SDFT's recovery is robust across random seeds, we now test whether this behavior generalizes to longer, more realistic task sequences. Using Qwen3.5-9B, we train four tasks sequentially: Science $\to$ Tooluse $\to$ Math $\to$ Code, with full SFT (not LoRA) at each stage. Full SFT is used here to simulate a more challenging continual learning scenario, where interference across tasks is stronger and more realistic for practical deployment. After each stage, we evaluate all four tasks; we then recover Science (task-1) from the final s4 model using the same four methods.

\begin{table}[htbp]
\centering
\small
\begin{tabular}{@{}lcccc@{}}
\toprule
After Stage & Science Acc\% & Tooluse Acc\% & Math Acc\% & Code Acc\% \\
\midrule
s1 science (science expert) & 72.6 & 46.4 & 65.3 & 67.3 \\
s2 tooluse & 61.3 & 69.1 & 62.0 & 9.3 \\
s3 math & 62.3 & 61.9 & 55.3 & 61.3 \\
s4 code (FINAL) & 59.8 & 67.0 & 51.3 & 58.0 \\
\bottomrule
\end{tabular}
\caption{Long task sequence in order (Science $\to$ Tooluse $\to$ Math $\to$ Code), using Qwen3.5-9B.}
\label{tab:longseq_forward}
\end{table}

\begin{table}[htbp]
\centering
\small
\begin{tabular}{@{}lcccc@{}}
\toprule
Method & Science Acc\% ($\Delta$ vs.\ Forgotten) & Tooluse Acc\% & Math Acc\% & Code Acc\% \\
\midrule
Forgotten (s4 FINAL) & 59.8 & 67.0 & 51.3 & 58.0 \\
SDFT & 75.1 (+15.3) & 69.1 & 56.7 & 64.0 \\
SFT & 71.8 (+12.0) & 68.0 & 54.0 & 62.7 \\
Fwd-KL KD & 75.9 (+16.1) & 69.1 & 53.3 & 62.7 \\
Replay (all four) & 72.8 (+13.0) & 67.0 & 54.0 & 59.3 \\
\bottomrule
\end{tabular}
\caption{Recovery of science after long task sequence across four methods, using Qwen3.5-9B.}
\label{tab:longseq_forward_recovery}
\end{table}

\paragraph{Results.}
Table~\ref{tab:longseq_forward} demonstrates the retention matrix after each stage. Science drops from 72.6 (after s1) to 59.8 (after s4), a total loss of 12.8 percentage points. Notably, this forgetting is not monotonic: Science partially recovers during Math training (+1.0 from s2 to s3) before declining again in Code training ($-$2.5 from s3 to s4), suggesting that intermediate tasks can have heterogeneous effects on previously learned capabilities. Code accuracy collapses to 9.3 after Tooluse training; we verified that this is due to format-level interference --- Tooluse training suppresses the model's ability to generate code in the right format. After Math training, Code rebounds to 61.3 (the reason for this rebound is not investigated in this work). The initial Code accuracy at s1 (67.3) reflects the model's pre-existing coding capability inherited from pre-training.

Table~\ref{tab:longseq_forward_recovery} presents the recovery results from the s4 final model. SDFT achieves Science recovery of 75.1\% (+15.3 from the forgotten state), outperforming SFT (71.8) and Replay (72.8), and is within 0.8 percentage points of fwd-KL KD (75.9). Critically, SDFT achieves the highest or tied-highest retention on all three non-target tasks: Tooluse (69.1, tied with fwd-KL KD), Math (56.7, highest), and Code (64.0, highest). In contrast, fwd-KL KD, while slightly higher on Science, shows lower retention on Math (53.3) and Code (62.7). Replay, despite requiring access to all four tasks' old data, achieves only 72.8 on Science and the worst Tooluse retention (67.0).

We also explored the reverse task order (Tooluse $\to$ Science $\to$ Code $\to$ Math), recovering Tooluse from the final model.

\begin{table}[htbp]
\centering
\small
\begin{tabular}{@{}lcccc@{}}
\toprule
After Stage & Tooluse Acc\% & Science Acc\% & Code Acc\% & Math Acc\% \\
\midrule
s1 tooluse (tooluse expert) & 71.1 & 43.0 & 33.3 & 69.3 \\
s2 science & 68.0 & 70.0 & 66.0 & 70.7 \\
s3 code & 66.0 & 70.4 & 57.3 & 52.0 \\
s4 math (FINAL) & 70.1 & 65.9 & 58.7 & 55.3 \\
\bottomrule
\end{tabular}
\caption{Long task sequence in reverse order (Tooluse $\to$ Science $\to$ Code $\to$ Math), using Qwen3.5-9B.}
\label{tab:longseq_reverse}
\end{table}

\begin{table}[htbp]
\centering
\small
\begin{tabular}{@{}lcccc@{}}
\toprule
Method & Tooluse Acc\% ($\Delta$ vs.\ Forgotten) & Science Acc\% & Code Acc\% & Math Acc\% \\
\midrule
Forgotten (s4 FINAL) & 70.1 & 65.9 & 58.7 & 55.3 \\
SDFT & 71.1 (+1.0) & 65.7 & 58.0 & 54.7 \\
SFT & 56.7 ($-$13.4) & 67.3 & 0.0 & 53.3 \\
Fwd-KL KD & 70.1 (+0.0) & 69.6 & 60.7 & 57.3 \\
Replay (all four) & 70.1 (+0.0) & 72.8 & 57.3 & 50.0 \\
\bottomrule
\end{tabular}
\caption{Recovery of science after long task sequence across four methods, using Qwen3.5-9B.}
\label{tab:longseq_reverse_recovery}
\end{table}

\paragraph{Results (Reverse Order).}
In the reverse order, Tooluse drops only slightly from 71.1 to 70.1 ($-$1.0) over the full sequence. SDFT fully recovers this small loss (70.1 $\to$ 71.1). While this recovery magnitude is modest, it reflects the fact that Tooluse itself experienced minimal forgetting over the full sequence. In contrast, SFT substantially degrades Tooluse from 70.1 to 56.7 ($-$13.4) and collapses on Code (0.0). This Code collapse is also due to the format-level interference observed in the forward order. Fwd-KL KD and Replay both achieve 70.1\% on Tooluse, matching the forgotten baseline, but Replay requires access to all four tasks' old data.

Taken together with the forward order results, these findings suggest that SDFT generalizes to different task orders and remains stable across longer, more complex task sequences.

\subsection{Testing the Manifold Alignment Hypothesis}
\label{sec:testing_hypothesis}

Having established that SDFT is an effective and robust recovery method across diverse degradation scenarios, we now turn to the mechanistic question: is this recovery consistently accompanied by manifold alignment? To investigate this, we conduct two sets of analyses: $\Delta$CKA--$\Delta$Acc association study and causal intervention experiment as below.

\paragraph{$\Delta$CKA--$\Delta$Acc Association Study.}
As defined in Section~\ref{sec:manifold_hypothesis}, the manifold alignment hypothesis predicts that effective recovery should be accompanied by an increase in CKA between the student and teacher (original Gemma-4-12B before pruning). To test this, we examine the relationship between $\Delta$CKA and $\Delta$Acc across below degradation scenarios.

\begin{table}[htbp]
\centering
\small
\begin{tabular}{@{}llcc@{}}
\toprule
Method & Task & Task Acc\% ($\Delta$ vs.\ Pruned Base) & CKA ($\Delta$ vs.\ Pruned Base) \\
\midrule
Pruned base & Tooluse & 47.42 & 0.757 \\
Pruned base & Science & 42.41 & 0.856 \\
SDFT & Tooluse & 56.70 (+9.28) & 0.742 ($-$0.015) \\
SDFT & Science & 45.76 (+3.53) & 0.851 ($-$0.005) \\
\bottomrule
\end{tabular}
\caption{$\Delta$CKA--$\Delta$Acc association study (Wanda 40\% Pruning) using Gemma-4-12B.}
\label{tab:cka_prune40}
\end{table}

\begin{table}[htbp]
\centering
\small
\begin{tabular}{@{}llcc@{}}
\toprule
Method & Task & Task Acc\% ($\Delta$ vs.\ Pruned Base) & CKA ($\Delta$ vs.\ Pruned Base) \\
\midrule
Pruned base & Tooluse & 51.55 & 0.666 \\
Pruned base & Science & 40.24 & 0.792 \\
SDFT & Tooluse & 60.82 (+9.27) & 0.541 ($-$0.125) \\
SDFT & Science & 46.55 (+6.31) & 0.799 (+0.007) \\
\bottomrule
\end{tabular}
\caption{$\Delta$CKA--$\Delta$Acc association study (Wanda 50\% Pruning) using Gemma-4-12B.}
\label{tab:cka_prune50}
\end{table}

\begin{table}[htbp]
\centering
\small
\begin{tabular}{@{}llcc@{}}
\toprule
Method & Task & Task Acc\% ($\Delta$ vs.\ 4bit) & CKA ($\Delta$ vs.\ 4bit) \\
\midrule
bf16 (reference) & Tooluse & --- & --- \\
bf16 (reference) & Science & --- & --- \\
NF4 4bit & Tooluse & 40.21 & 0.774 \\
NF4 4bit & Science & 51.08 & 0.859 \\
SDFT & Tooluse & 52.58 (+12.37) & 0.785 (+0.011) \\
SDFT & Science & 59.17 (+8.09) & 0.868 (+0.009) \\
\bottomrule
\end{tabular}
\caption{$\Delta$CKA--$\Delta$Acc association study (NF4 quantization) using Gemma-4-12B.}
\label{tab:cka_nf4}
\end{table}

\paragraph{Results.}
Tables~\ref{tab:cka_prune40} and~\ref{tab:cka_prune50} present the CKA changes associated with SDFT recovery under Wanda 40\% and 50\% pruning. Under 40\% pruning (Table~\ref{tab:cka_prune40}), SDFT recovers Tooluse by +9.28 and Science by +3.53, yet CKA decreases on both tasks ($-$0.015 on Tooluse, $-$0.005 on Science). Under 50\% pruning (Table~\ref{tab:cka_prune50}), SDFT recovers Tooluse by +9.27 and Science by +6.31; CKA decreases on Tooluse ($-$0.125) and shows only a negligible increase on Science (+0.007). In both cases, performance recovery occurs without corresponding CKA increase, contradicting the manifold alignment hypothesis in this scenario.

The pruning results (Tables~\ref{tab:cka_prune40}--\ref{tab:cka_prune50}) provide a direct counterexample to the manifold alignment hypothesis: SDFT recovers performance without increasing CKA. The quantization results (Table~\ref{tab:cka_nf4}), by contrast, show the only setting where a consistent positive $\Delta$CKA--$\Delta$Acc association is observed.

We did not separately test the $\Delta$CKA--$\Delta$Acc association in the forgetting scenario, as the pruning results already established that the manifold alignment hypothesis is not universal. Instead, we directly probe the forgetting scenario in the causal intervention experiment below to assess whether any functional role of manifold alignment is detectable when no prior correlational evidence exists.

\paragraph{Causal Intervention Experiment.}
To probe whether this association reflects a functional role, we conduct a causal intervention experiment using Qwen3.5-9B, where we add a regularization term $\lambda \cdot \mathrm{CKA}(\text{student}, \text{teacher})$ to the SDFT loss to explicitly penalize manifold alignment ($\lambda \in \{0, 0.1, 0.3\}$). The results across three random seeds are shown in below tables.

\begin{table}[htbp]
\centering
\small
\resizebox{\textwidth}{!}{%
\begin{tabular}{@{}llcccc@{}}
\toprule
Method & $\lambda$ & Science Acc\% ($\Delta$ vs.\ forgotten) & CKA (Train, start$\to$final) & CKA (Eval) & Logit-KL (Train) \\
\midrule
Science expert & --- & 60.75 & --- & --- & --- \\
Forgotten & --- & 59.96 & --- & 0.985 & --- \\
SDFT & 0 (No penalty) & 63.31 (+3.35) & --- & 0.987 & --- \\
SDFT & 0.1 & 64.10 (+4.14) & 0.993 $\to$ 0.177 & 0.125 & 0.028 \\
SDFT & 0.3 & 59.57 ($-$0.39) & 0.993 $\to$ 0.132 & 0.146 & 0.033 \\
\bottomrule
\end{tabular}%
}
\caption{Causal intervention experiment with pipeline (Science $\to$ Tooluse $\to$ Recovery), random seed 42, using Qwen3.5-9B.}
\label{tab:causal_s42}
\end{table}

\begin{table}[htbp]
\centering
\small
\resizebox{\textwidth}{!}{%
\begin{tabular}{@{}llcccc@{}}
\toprule
Method & $\lambda$ & Science Acc\% ($\Delta$ vs.\ forgotten) & CKA (Train, start$\to$final) & CKA (Eval) & Logit-KL (Train) \\
\midrule
Science expert & --- & 61.14 & --- & --- & --- \\
Forgotten & --- & 55.23 & --- & 0.984 & \\
SDFT & 0 (No penalty) & 62.52 (+7.29) & --- & 0.990 & \\
SDFT & 0.1 & 61.14 (+5.91) & 0.993 $\to$ 0.181 & 0.231 & 0.028 \\
SDFT & 0.3 & 58.78 (+3.55) & 0.993 $\to$ 0.127 & 0.174 & 0.032 \\
\bottomrule
\end{tabular}%
}
\caption{Causal intervention experiment with pipeline (Science $\to$ Tooluse $\to$ Recovery), random seed 43, using Qwen3.5-9B.}
\label{tab:causal_s43}
\end{table}

\begin{table}[htbp]
\centering
\small
\resizebox{\textwidth}{!}{%
\begin{tabular}{@{}llcccc@{}}
\toprule
Method & $\lambda$ & Science Acc\% ($\Delta$ vs.\ forgotten) & CKA (Train, start$\to$final) & CKA (Eval) & Logit-KL (Train) \\
\midrule
Science expert & --- & 60.36 & --- & --- & --- \\
Forgotten & --- & 52.86 & --- & 0.983 & --- \\
SDFT & 0 (No penalty) & 61.74 (+8.88) & --- & 0.989 & --- \\
SDFT & 0.1 & 59.96 (+7.10) & 0.993 $\to$ 0.161 & 0.199 & 0.029 \\
SDFT & 0.3 & 58.38 (+5.52) & 0.993 $\to$ 0.123 & 0.108 & 0.034 \\
\bottomrule
\end{tabular}%
}
\caption{Causal intervention experiment with pipeline (Science $\to$ Tooluse $\to$ Recovery), random seed 44, using Qwen3.5-9B.}
\label{tab:causal_s44}
\end{table}

\begin{table}[htbp]
\centering
\small
\begin{tabular}{@{}lcc@{}}
\toprule
$\lambda$ & Science Acc\% $\Delta$ (mean $\pm$ std) & CKA (Eval) (mean $\pm$ std) \\
\midrule
0 & +6.51 $\pm$ 2.32 & 0.984 $\pm$ 0.001 \\
1 & +5.72 $\pm$ 1.22 & 0.185 $\pm$ 0.044 \\
2 & +2.89 $\pm$ 2.46 & 0.143 $\pm$ 0.027 \\
\bottomrule
\end{tabular}
\caption{Causal intervention results averaged over 3 random seeds in the forgetting-recovery pipeline (Qwen3.5-9B). Science recovery ($\Delta$, mean $\pm$ std over seeds) and CKA (Eval) across different penalty strengths $\lambda$. $\lambda$=0 corresponds to standard SDFT without penalty.}
\label{tab:causal_forget_avg}
\end{table}

\paragraph{Causal Intervention Results (Tables~\ref{tab:causal_s42}--\ref{tab:causal_forget_avg}).}
The penalty effectively suppresses CKA from 0.984 to 0.185 ($\lambda$=0.1) and further to 0.143 ($\lambda$=0.3) across all three random seeds, confirming the intervention operates as intended. At $\lambda$=0.1, Science recovery shows a modest decrease from +6.51 to +5.72 ($\Delta$ = $-$0.79), with mixed per-seed behavior (s42: +4.14, s43: +5.91, s44: +7.10). At $\lambda$=0.3, Science recovery consistently weakens to +2.89 ($\Delta$ = $-$2.83 from $\lambda$=0.1; $\Delta$ = $-$3.62 from $\lambda$=0), with all three seeds showing lower recovery than at $\lambda$=0.1 (s42: $-$4.53, s43: $-$2.36, s44: $-$1.58). Logit-KL (0.028--0.034) remains stable across all conditions.

While individual seeds show some variability --- particularly at intermediate penalty strength --- the summary statistics across three random seeds reveal a clear pattern: stronger CKA suppression correlates with weaker recovery ($\lambda$=0: +6.51 $\to$ $\lambda$=0.1: +5.72 $\to$ $\lambda$=0.3: +2.89), and the causal intervention consistently reduces recovery when alignment is sufficiently suppressed, supporting the functional relevance of manifold realignment in the quantization scenario.

\begin{table}[htbp]
\centering
\small
\resizebox{\textwidth}{!}{%
\begin{tabular}{@{}lll@{}}
\toprule
Seed & Science Acc\% (bf16/NF4/$\lambda$=0/0.1/0.3) & CKA (Eval) (NF4/$\lambda$=0/0.1/0.3) \\
\midrule
42 & 60.75 / 54.24 / 64.30 / 62.52 / 60.75 & 0.987 / 0.989 / 0.409 / 0.391 \\
43 & 61.14 / 54.24 / 61.93 / 61.93 / 59.37 & 0.988 / 0.988 / 0.371 / 0.288 \\
44 & 60.36 / 53.65 / 62.52 / 58.19 / 60.36 & 0.986 / 0.988 / 0.336 / 0.323 \\
\bottomrule
\end{tabular}%
}
\caption{Causal intervention experiment in NF4 quantization scenario over 3 random seeds (Qwen3.5-9B).}
\label{tab:causal_nf4_seeds}
\end{table}

\begin{table}[htbp]
\centering
\small
\begin{tabular}{@{}lcc@{}}
\toprule
$\lambda$ & Science Acc\% $\Delta$ (mean $\pm$ std) & CKA (Eval) (mean $\pm$ std) \\
\midrule
0 & +8.28 $\pm$ 1.45 & 0.988 $\pm$ 0.000 \\
1 & +6.84 $\pm$ 1.21 & 0.372 $\pm$ 0.030 \\
2 & +6.71 $\pm$ 1.77 & 0.334 $\pm$ 0.043 \\
\bottomrule
\end{tabular}
\caption{Causal intervention results averaged over 3 random seeds in NF4 quantization (Qwen3.5-9B). Science recovery ($\Delta$, mean $\pm$ std over seeds) and CKA (Eval) across different penalty strengths $\lambda$. $\lambda$=0 corresponds to standard SDFT without penalty.}
\label{tab:causal_nf4_avg}
\end{table}

\paragraph{Quantization Causal Intervention (Tables~\ref{tab:causal_nf4_seeds}--\ref{tab:causal_nf4_avg}).}
The penalty effectively suppresses CKA from 0.988 to 0.372 ($\lambda$=0.1) and further to 0.334 ($\lambda$=0.3) across all three seeds. Science recovery decreases from +8.28 to +6.84 at $\lambda$=0.1 ($\Delta$ = $-$1.44), and to +6.71 at $\lambda$=0.3 ($\Delta$ = $-$0.13 from $\lambda$=0.1). All three seeds show lower Science recovery at $\lambda$=0.3 than at $\lambda$=0 (s42: 64.30$\to$60.75, s43: 61.93$\to$59.37, s44: 62.52$\to$60.36), confirming the direction of the effect. While individual seeds show higher variability comparing to forgetting-recovery, the consistent direction across all three seeds provides supporting evidence that manifold realignment plays a functional role in quantization recovery via SDFT.

\begin{table}[htbp]
\centering
\small
\resizebox{\textwidth}{!}{%
\begin{tabular}{@{}llccccc@{}}
\toprule
Task & $\lambda$ & Task Acc\% & $\Delta$ vs.\ 4bit & Logit-KL (Train) & CKA (Train, start$\to$final) & CKA (Eval) \\
\midrule
Tooluse (NF4 4bit) & --- & 40.21 & --- & --- & --- & --- \\
Tooluse & 0 (No penalty) & 52.58 & +12.37 & 0.080 & 0.907 $\to$ 0.907 (*) & 0.785 \\
Tooluse & 0.1 & 50.52 & +10.31 & 0.082 & 0.907 $\to$ 0.123 & 0.399 \\
Tooluse & 0.3 & 49.48 & +9.27 & 0.085 & 0.907 $\to$ 0.093 & 0.460 \\
\midrule
Science (NF4 4bit) & --- & 51.08 & --- & --- & --- & --- \\
Science & 0 (No penalty) & 59.17 & +8.09 & 0.103 & 0.944 $\to$ 0.944 (*) & 0.868 \\
Science & 0.1 & 56.21 & +5.13 & 0.107 & 0.944 $\to$ 0.133 & 0.139 \\
Science & 0.3 & 53.25 & +2.17 & 0.111 & 0.944 $\to$ 0.081 & 0.122 \\
\bottomrule
\end{tabular}%
}
\caption{Causal intervention experiment in NF4 quantization scenario (Gemma-4-12B).}
\label{tab:causal_nf4_gemma}
\end{table}

\paragraph{Causal Intervention (Quantization, Gemma-4-12B, Single Run).}
Table~\ref{tab:causal_nf4_gemma} presents the causal intervention results in the NF4 quantization scenario using Gemma-4-12B. The penalty effectively suppresses CKA from $\sim$0.9 to 0.1--0.4 across both tasks. Science recovery decreases monotonically with penalty strength: +8.09 ($\lambda$=0) $\to$ +5.13 ($\lambda$=0.1) $\to$ +2.17 ($\lambda$=0.3). Tooluse recovery shows a similar trend: +12.37 $\to$ +10.31 $\to$ +9.27. Logit-KL remains stable across all $\lambda$ values (0.080--0.085 for Tooluse, 0.103--0.111 for Science), confirming that the penalty specifically targets manifold alignment without disrupting logits-level distillation. This single-run result is consistent with the multi-seed quantization experiments on Qwen3.5-9B (Tables~\ref{tab:causal_nf4_seeds}--\ref{tab:causal_nf4_avg}), providing cross-model corroboration of the functional role of manifold alignment in quantization recovery.

\section{Discussion}
\label{sec:discussion}

\subsection{Summary of Findings}

\textbf{First,} SDFT is an effective and robust recovery method across catastrophic forgetting, quantization, and pruning. Baseline comparisons, multi-seed experiments, and long task sequences (forward and reverse orders) consistently confirm its trade-off-free recovery.

\textbf{Second,} we tested the manifold alignment hypothesis. In the quantization and forgetting scenarios, SDFT recovery is consistently accompanied by CKA increase (Tables~\ref{tab:cka_nf4}--\ref{tab:causal_s44}). In the pruning scenario, however, we observe counterexamples: SDFT recovers performance while CKA decreases or stagnates (Tables~\ref{tab:cka_prune40}--\ref{tab:cka_prune50}), indicating that manifold alignment is not a general requirement for recovery.

\textbf{Third,} at the aggregate level, suppressing CKA weakens recovery in both quantization and forgetting, as shown by the dose-dependent trend across three random seeds (Tables~\ref{tab:causal_forget_avg},\ref{tab:causal_nf4_avg}). We note that individual seeds show some variability, but the statistical trend is consistent across both scenarios.

\textbf{Fourth,} the pruning results provide a clear boundary condition: SDFT can recover without manifold alignment. The mechanism of SDFT recovery is not reducible to a single geometric principle, and varies across degradation types.

\textbf{Together,} these findings establish SDFT as a broadly effective recovery method, while delineating the boundary of the manifold alignment hypothesis --- it holds in quantization and forgetting, but fails in pruning.

\subsection{Limitations}

\textbf{CKA is dataset-dependent.} As discussed in Section~\ref{sec:manifold_hypothesis}, CKA values are computed from activations produced by a specific input dataset, and their absolute values are not comparable across different datasets. Our analysis relies on relative changes ($\Delta$CKA) within a fixed dataset throughout the recovery process.

\textbf{Causal intervention results are based on aggregate trends.} While the dose-dependent pattern is consistent across three random seeds in both forgetting and quantization, individual seeds show some variability. The conclusion that suppressing CKA weakens recovery holds at the statistical level, not for every single run.

\textbf{Our analysis focuses on the last hidden layer.} We chose this layer because it is most directly connected to downstream performance, but layer-wise dynamics may reveal additional insights. Future work could explore whether different layers exhibit different alignment patterns across degradation types.

\textbf{Generalizability to other architectures and modalities remains to be tested.} While we have included Qwen and Gemma as cross-family validation, broader coverage including larger models and other modalities (e.g., vision-language models) would strengthen the conclusions.

\subsection{Future Work}

Several directions remain open.

\textbf{First,} the pruning scenario presents a clear counterexample where SDFT recovers performance without CKA increase. Understanding what mechanism drives recovery in this setting and whether it generalizes to other structural degradations is a natural next step.

\textbf{Second,} our CKA analysis focused on the last hidden layer; layer-wise or block-wise analysis could reveal whether alignment patterns differ across layers and whether certain layers are more critical in specific scenarios.

\textbf{Third,} while we have validated our findings on Qwen and Gemma across multiple tasks, extending the analysis to larger models (e.g., 70B+) and other modalities (e.g., vision-language models) would further test the generality of our conclusions.

\textbf{Finally,} the differential sensitivity of quantization and forgetting to CKA suppression suggests that different degradation types may engage different recovery pathways; characterizing these pathways more precisely is an important direction for future research.

\section{Conclusion}
\label{sec:conclusion}

In this work, we have systematically evaluated SDFT as a recovery method for LLMs across three degradation scenarios --- catastrophic forgetting, quantization, and pruning. Through baseline comparisons, multi-seed experiments, and long task sequences in both forward and reverse orders, we establish that SDFT consistently and robustly recovers degraded capabilities while preserving others.

We then tested the manifold alignment hypothesis --- the idea that SDFT recovery operates through the realignment of hidden representations toward a teacher model, as measured by CKA. Our results reveal a differentiated picture. In quantization and forgetting, recovery is consistently accompanied by CKA increase, and causal intervention experiments show that suppressing CKA weakens recovery at the aggregate level. In pruning, however, we observe clear counterexamples: SDFT recovers performance while CKA decreases or stagnates. The pruning scenario thus serves as a boundary condition: manifold alignment is not a universal requirement for SDFT recovery.

These findings demonstrate that SDFT is a broadly effective and robust recovery method, while the mechanism underlying its effectiveness varies across degradation types. The manifold alignment hypothesis holds in quantization and forgetting, but fails in pruning. This boundary condition is a key contribution of our work --- it delineates where the geometric interpretation applies and where it does not, providing a constraint for future theoretical accounts of SDFT recovery.

For practitioners, SDFT remains a reliable recovery method across all three settings. For researchers, our results offer a clear direction: future work should investigate what alternative mechanisms enable SDFT to recover performance in scenarios where manifold alignment is absent.

\subsubsection*{Broader Impact Statement}
This work proposes a recovery mechanism for LLM performance degradation. While the framework itself is a general-purpose tool for improving model quality, we note that enhanced LLM capabilities could amplify both beneficial and harmful applications. We encourage practitioners to apply responsible deployment practices when using recovered models in production systems.

\bibliography{example_paper}

@article{shenfeld2026sdft,
  title={Self-Distillation Enables Continual Learning},
  author={Shenfeld, Idan and Damani, Mehul and H{\"u}botter, Jonas and Agrawal, Pulkit},
  journal={arXiv preprint arXiv:2601.19897},
  year={2026}
}

@article{hui2024qwen2,
  title={Qwen2.5 Technical Report},
  author={Qwen, Team and Yang, An and Yang, Baosong and Zhang, Beichen and Hui, Binyuan and Zheng, Bo and Yu, Bowen and Li, Chengyuan and Liu, Dayiheng and Huang, Fei and others},
  journal={arXiv preprint arXiv:2412.15115},
  year={2024}
}

@article{de2021continual,
  title={A continual learning survey: Defying forgetting in classification tasks},
  author={De Lange, Matthias and Aljundi, Rahaf and Masana, Marc and Parisot, Sarah and Jia, Xu and Leonardis, Ale{\v{s}} and Slabaugh, Gregory and Tuytelaars, Tinne},
  journal={IEEE transactions on pattern analysis and machine intelligence},
  volume={44},
  number={7},
  pages={3366--3385},
  year={2021},
  publisher={IEEE}
}

@misc{eval-harness,
  author       = {Gao, Leo and Tow, Jonathan and Abbasi, Baber and Biderman, Stella and Black, Sid and DiPofi, Anthony and Foster, Charles and Golding, Laurence and Hsu, Jeffrey and Le Noac'h, Alain and Li, Haonan and McDonell, Kyle and Muennighoff, Niklas and Ociepa, Chris and Phang, Jason and Reynolds, Laria and Schoelkopf, Hailey and Skowron, Aviya and Sutawika, Lintang and Tang, Eric and Thite, Anish and Wang, Ben and Wang, Kevin and Zou, Andy},
  title        = {The Language Model Evaluation Harness},
  month        = 07,
  year         = 2024,
  publisher    = {Zenodo},
  version      = {v0.4.3},
  doi          = {10.5281/zenodo.12608602},
  url          = {https://zenodo.org/records/12608602}
}

@article{hinton2015distilling,
  title={Distilling the knowledge in a neural network},
  author={Hinton, Geoffrey and Vinyals, Oriol and Dean, Jeff},
  journal={arXiv preprint arXiv:1503.02531},
  year={2015}
}

@article{kirkpatrick2017overcoming,
  title={Overcoming catastrophic forgetting in neural networks},
  author={Kirkpatrick, James and Pascanu, Razvan and Rabinowitz, Neil and Veness, Joel and Desjardins, Guillaume and Rusu, Andrei A and Milan, Kieran and Quan, John and Ramalho, Tiago and Grabska-Barwinska, Agnieszka and others},
  journal={Proceedings of the national academy of sciences},
  volume={114},
  number={13},
  pages={3521--3526},
  year={2017},
  publisher={National Academy of Sciences}
}

@article{li2017learning,
  title={Learning without forgetting},
  author={Li, Zhizhong and Hoiem, Derek},
  journal={IEEE transactions on pattern analysis and machine intelligence},
  volume={40},
  number={12},
  pages={2935--2947},
  year={2017},
  publisher={IEEE}
}

@article{dettmers2023qlora,
  title={QLoRA: Efficient Finetuning of Quantized {LLMs}},
  author={Dettmers, Tim and Pagnoni, Artidoro and Holtzman, Ari and Zettlemoyer, Luke},
  journal={Advances in Neural Information Processing Systems},
  volume={36},
  year={2023}
}

@article{gretton2005measuring,
  title={Measuring Statistical Dependence with Hilbert-Schmidt Norms},
  author={Gretton, Arthur and Bousquet, Olivier and Smola, Alex and Sch{\"o}lkopf, Bernhard},
  journal={Algorithmic Learning Theory},
  pages={63--77},
  year={2005},
  publisher={Springer}
}

@inproceedings{kornblith2019similarity,
  title={Similarity of Neural Network Representations Revisited},
  author={Kornblith, Simon and Norouzi, Mohammad and Lee, Honglak and Hinton, Geoffrey},
  booktitle={International Conference on Machine Learning},
  pages={3519--3529},
  year={2019},
  organization={PMLR}
}

@inproceedings{frantar2022gptq,
  title={{GPTQ}: Accurate Post-Training Quantization for Generative Pre-trained Transformers},
  author={Frantar, Elias and Ashkboos, Saleh and Hoefler, Torsten and Alistarh, Dan},
  booktitle={International Conference on Learning Representations},
  year={2023}
}

@inproceedings{furlanello2018born,
  title={Born Again Neural Networks},
  author={Furlanello, Tommaso and Lipton, Zachary C. and Tschannen, Michael and Itti, Laurent and Anandkumar, Anima},
  booktitle={Proceedings of the 35th International Conference on Machine Learning (ICML 2018)},
  pages={1607--1616},
  year={2018}
}

@article{ma2024llmpruner,
  title={{LLM-Pruner}: On the Structural Pruning of Large Language Models},
  author={Ma, Xinyin and Fang, Gongfan and Wang, Xinchao},
  journal={Advances in Neural Information Processing Systems},
  volume={36},
  year={2024}
}

@article{rusu2016progressive,
  title={Progressive Neural Networks},
  author={Rusu, Andrei A. and Rabinowitz, Neil C. and Desjardins, Guillaume and Soyer, Hubert and Kirkpatrick, James and Kavukcuoglu, Koray and Pascanu, Razvan and Hadsell, Raia},
  journal={arXiv preprint arXiv:1606.04671},
  year={2016}
}

@article{thangarasa2025self,
  title={Self-Data Distillation for Recovering Quality in Pruned Large Language Models},
  author={Thangarasa, Vithursan and Venkatesh, Ganesh and Lasby, Mike and Sinnadurai, Nish and Lie, Sean},
  journal={arXiv preprint arXiv:2410.09982},
  year={2025}
}

@misc{qwen2026qwen35,
  title={{Qwen3.5}: Towards Native Multimodal Agents},
  author={Qwen, Team and others},
  year={2026},
  howpublished={\url{https://qwen.ai/blog?id=qwen3.5}}
}

@article{gemmateam2026gemma4,
  title={Gemma 4 Technical Report},
  author={{Gemma Team}},
  journal={arXiv preprint arXiv:2607.02770},
  year={2026}
}
\bibliographystyle{tmlr}

\appendix

\section{Training and Implementation Details}
\label{app:training}

This appendix details the training configuration used across our experiments. Unless otherwise noted, the same recipe is applied to every method and scenario.

\subsection{Models and Degradation Operators}
We use Qwen2.5-3B-Instruct \citep{hui2024qwen2}, Qwen3.5-9B \citep{qwen2026qwen35}, and Gemma-4-12B \citep{gemmateam2026gemma4} as base models, selected per experiment as indicated in the corresponding table captions. Three degradation operators are considered: (i)~\emph{catastrophic forgetting}, induced by sequential fine-tuning on new domains (LoRA or full fine-tuning); (ii)~\emph{quantization}, using 4-bit NF4 via bitsandbytes; and (iii)~\emph{pruning}, using Wanda structured pruning at 40\% and 50\% sparsity.

\subsection{Tasks and Datasets}
Recovery and expert tasks use fixed on-disk train/eval splits: Science ($2{,}674$ / $507$), Tooluse ($4{,}046$ / $97$), Math (GSM8K; $4{,}000$ / $500$), and Code ($4{,}000$ / $500$; used in the long-task-sequence experiments of Section~\ref{sec:experiments}). General-capability assessment additionally draws on MMLU, Winogrande, ARC-Challenge, HellaSwag, GSM8K, and TruthfulQA. Because the evaluation sets are fixed on disk and decoding is deterministic (Section~\ref{app:eval}), evaluation contributes no measurement noise; all cross-seed variance reported in the paper originates from training stochasticity.

\subsection{Parameter-Efficient Fine-Tuning}
All parameter-efficient fine-tuning uses LoRA with rank $r=16$, scaling $\alpha=32$, and dropout $0.05$, applied to the query, key, value, output, gate, up, and down projection matrices (\texttt{q/k/v/o/gate/up/down}). The long-task-sequence experiment (Section~\ref{sec:experiments}) instead uses full fine-tuning to simulate a more challenging continual-learning setting.

\subsection{Optimization}
We train in bf16 with a peak learning rate of $5\times10^{-5}$ under a cosine schedule with $10\%$ linear warmup, for $2$ epochs. The effective batch size is $32$, realized as a per-device batch size of $1$ with gradient accumulation over $32$ steps; at this configuration two epochs correspond to approximately $166$ optimizer steps on Science and $252$ on Tooluse. LoRA runs use AdamW; full-parameter fine-tuning uses paged 8-bit AdamW with gradient checkpointing and loads the frozen teacher in 8-bit to fit memory. Multi-seed experiments use training seeds $42$, $43$, and $44$ (default $42$), which control data shuffling and initialization.

\subsection{SDFT Objective}
SDFT performs on-policy self-distillation: the student generates completions (prompt and completion lengths each capped at $1{,}024$ tokens; teacher generation optionally accelerated with vLLM), which are then scored against a frozen reference (Section~\ref{sec:recovery_framework}). The distillation weight $\alpha$ selects the divergence direction: $\alpha=0$ recovers forward KL, $\alpha\in(0,1)$ yields a Jensen--Shannon-style objective, and $\alpha=1$ (our default) gives the reverse-KL objective used throughout. The reference (teacher) is the bf16 original model for compression recovery and the pre-degradation checkpoint for forgetting recovery. A small reference-model mixup coefficient of $0.01$ is used by default. For compression recovery, each method is trained independently on the Tooluse and Science domains (separate runs).

\subsection{Baselines}
We compare against: \textbf{SFT} (standard LoRA fine-tuning); \textbf{forward-KL KD} (knowledge distillation with a forward-KL objective); \textbf{EWC} \citep{kirkpatrick2017overcoming}, anchored to the degraded model with penalty $\lambda=10^{4}$; \textbf{LwF} \citep{li2017learning}, anchored to the degraded model with $\lambda=1.0$; and \textbf{Replay}, which jointly trains on the target-recovery data together with retained data from the other task(s).

\subsection{CKA Computation}
Centered Kernel Alignment is computed from the last-layer hidden states, restricted to non-padding tokens, on a fixed evaluation set (activation extraction capped at $512$ tokens). We use a linear kernel $K = X X^\top$ and global (double) centering $H = I - \frac{1}{n}\mathbf{1}\mathbf{1}^\top$, so that $\mathrm{HSIC}(A,B) = \mathrm{tr}(K_A H K_B H)/(n-1)^2$ and $\mathrm{CKA}(A,B) = \mathrm{HSIC}(A,B)/\sqrt{\mathrm{HSIC}(A,A)\,\mathrm{HSIC}(B,B)}$; the $(n-1)^{-2}$ factor cancels in the ratio (Section~\ref{sec:manifold_hypothesis}). Because absolute CKA values depend on the evaluation dataset, all analyses report relative changes ($\Delta$CKA) within a fixed dataset across recovery stages.

\subsection{Causal Intervention}
To probe the functional role of manifold alignment, we optimize $\mathcal{L} = \mathcal{L}_{\mathrm{SDFT}} + \lambda \cdot \mathrm{CKA}(H_{\mathrm{student}}, H_{\mathrm{teacher}})$, penalizing rather than encouraging alignment, with $\lambda \in \{0, 0.1, 0.3\}$. The penalty CKA is computed as above on the last-layer hidden states; $\lambda=0$ is bit-identical to plain SDFT and serves as the control. Multi-seed results use random seeds $42$, $43$, and $44$.

\subsection{Evaluation}
\label{app:eval}
Task-specific accuracy is reported on Science, Tooluse, Math, and Code. General capability is evaluated on MMLU, Winogrande, ARC-Challenge, HellaSwag, GSM8K, and TruthfulQA using the LM Evaluation Harness \citep{eval-harness}. All decoding is deterministic greedy (\texttt{do\_sample=False}); multiple-choice items are scored by exact match on the predicted answer letter.

\end{document}